%% file: main.tex
\documentclass[a4paper,table,16pt]{extarticle}
\usepackage{booktabs}
\usepackage{colortbl} % 用于彩色字体
\usepackage{xcolor}   % 定义颜色
\usepackage[table]{xcolor}
\definecolor{pelicanmain}{HTML}{E2E8F0}  % Soft professional gray-blue
\definecolor{pelicanlight}{HTML}{F8F9FA} % Ultra light gray
\usepackage{xspace}
\usepackage{tabularx} % 自适应表格宽度
\usepackage{booktabs} % 美观的横线
\usepackage{graphicx}
\usepackage{natbib}
\usepackage{amssymb}
\usepackage{arpafvg}

\usepackage{multirow}  % 导言区添加
\usepackage{colortbl}

\usepackage{helvet}
\usepackage{helvet}
\usepackage{titlesec}
\usepackage{xcolor}
\usepackage{CJKutf8}

\definecolor{titleorange}{RGB}{236,120,47}
\definecolor{rowgreen}{RGB}{234,243,222}
\definecolor{rowblue}{RGB}{230,241,251}
\definecolor{rowpurple}{RGB}{238,237,254}

\usepackage{helvet}
\usepackage{titlesec}
\usepackage{xcolor}

\definecolor{titleorange}{RGB}{236,120,47}

\titleformat{\section}
{\normalfont\fontfamily{phv}\fontseries{bx}\selectfont\Large\color{titleorange}}
{\thesection.}{0.5em}{}

\titleformat{\subsection}
{\normalfont\fontfamily{phv}\fontseries{bx}\selectfont\normalsize\color{black}}
{\thesubsection.}{0.5em}{}

\titleformat{\subsubsection}
{\normalfont\fontfamily{phv}\fontseries{bx}\selectfont\normalsize\color{black}}
{\thesubsubsection.}{0.5em}{}

\usepackage[most]{tcolorbox}
\definecolor{mylightgreen}{HTML}{D7FFD7}
\definecolor{myred}{RGB}{255,0,0}
\definecolor{modiblue}{RGB}{0,40,161}
\usepackage{lipsum}      % 用于生成示例文本，您可以删除这一行
\usepackage{booktabs} % 提供漂亮的表格线 \toprule, \midrule, \bottomrule
\usepackage{multirow} % 支持多行单元格
\usepackage{caption} % 用于表格标题
\usepackage{makecell}
\usepackage{adjustbox} % 导言区添加
\newtcolorbox{takeawaybox}{
  colback=gray!10,  % 背景颜色，10%的灰色
  colframe=gray!75, % 边框颜色
  boxrule=0.5pt,    % 边框线宽
  arc=2mm,          % 圆角弧度，如果想要直角，可以用 sharp corners
  halign=center,    % 内容水平居中
  valign=center,    % 内容垂直居中（对于单行文本效果不明显）
  boxsep=3pt,       % 文本和边框的间距
  fontupper=\small\bfseries\sffamily, % 文本样式：大号、加粗、无衬线字体
  sharp corners,    % 使用直角，更符合学术风格
}
\newtcolorbox{casebox}[1][]{
    enhanced,
    colback=black!5!white, % 背景颜色：5%的黑色（非常浅的灰色）
    colframe=black!60!white, % 边框颜色：60%的黑色（深灰色）
    boxrule=0.8pt,           % 边框线宽
    arc=2mm,                 % 边框圆角弧度
    fonttitle=\bfseries\sffamily, % 标题字体：加粗、无衬线
    title=#1,                % 将环境的参数作为盒子的标题
    left=6pt, right=6pt, top=6pt, bottom=6pt, % 内边距
    sharp corners,
    breakable,               % 允许盒子内容跨页
    #1 % 允许在调用时传入额外的 tcolorbox 参数
}

\usepackage{float} % 支持 [H] 放置
\usepackage{epigraph}
\begin{document}
\pagecolor{white}

\makeatletter

\title{\textcolor{titlecolor}{Pelican-Unify 1.0: \\ A Unified Embodied Intelligence Model (UEI) \\ for Understanding, Reasoning, Imagination and Action}}

% \date{\today}

\makeatother

\author{
	Beijing Innovation Center of Humanoid Robotics (X-Humanoid) \\
	% \textit{PelicanVLM.github.io} \\
    \textbf{WFM System Group} \\
    \{vito.dai,jian.tang,jason.ju\}@x-humanoid.com
    }
% \date{\today}

\maketitle

\section*{\textbf{Abstract}}
\vspace{-0.3em}

\begin{center}
\begin{minipage}{0.86\textwidth}
\centering
\small\itshape
``The soul never thinks without an image.'' --- Aristotle\\[-0.1em]
``My thinking is first and last and always for the sake of my doing.'' --- William James\\[0.1em]
\begin{CJK*}{UTF8}{gbsn}
\textit{博学之，审问之，慎思之，明辨之，笃行之。} ---《礼记·中庸》
\end{CJK*}

\end{minipage}
\end{center}

\vspace{0.25em}

\noindent
\emph{We argue that foundation models for embodied intelligence should not be built upon fragmented capabilities.} Rather, understanding, reasoning, imagination, and action should be regarded as interdependent dimensions in a single adaptive intelligence loop. The advancement of an embodied agent in the physical world does not arise from possessing independent vision model, language model, world model, and action policy, but from its capability to integrate world understanding, task reasoning, future imagination, and action execution within a latent world space that can be aligned, abstracted, planned, and refined. We therefore advocate a unified paradigm for building embodied intelligence models of physical intelligence.

By “unified,” we do not refer to simple concatenation of multiple expert networks at their outputs, nor to the assembly of independently optimized modules into a sequential pipeline. 
\emph{We mean structurally shared representations, mutually constrained conditions, and co-evolution through a common training process.} Concretely, a truly unified model should exhibit three key properties: \textbf{unified understanding}, which embeds scenes, instructions, action histories, and visual contexts into a shared semantic space for a holistic understanding of what the agent sees, what it needs to accomplish, what it has done, and what state the world is in; \textbf{unified reasoning}, which turns reasoning into a language-grounded, supervisable process over task intent, action choices, and future consequences, rather than a linguistic monologue detached from action and imagination; and \textbf{unified generation}, which jointly reads out future imagination and low-level action from the same diffusion decoding process conditioned on the same reasoning latent, so actions are shaped by imagined consequences, imagination is framed by task reasoning, and reasoning is constrained by what can be imagined and executed. \emph{In this way, understanding, reasoning, imagination, and action become a collaborative system that exchanges gradients, corrects itself, and strengthens itself within a single training loop, rather than separate stages in a pipeline.}

Based on this paradigm, we present \textbf{Pelican-Unify 1.0}, the first embodied foundation model trained according to the principle of unification. Pelican-Unify 1.0 uses a single VLM as a unified understanding module, mapping scenes, instructions, visual contexts, and action histories into a shared semantic space. The same VLM also serves as a unified reasoning module, autoregressively producing task-, action-, and future-oriented chains of thought in a single forward pass and projecting the final hidden state into a dense latent variable. A Unified Future Generator (UFG) then conditions on this latent variable and jointly generates future videos and future actions through two modality-specific output heads within the same denoising process. The language, video, and action losses are all backpropagated into the shared representation, enabling the model to jointly optimize understanding, reasoning, imagination, and action during training, rather than training three isolated expert systems.

Experiments demonstrate that unification does not imply compromise. With a single checkpoint, Pelican-Unify 1.0 achieves strong performance across all three capabilities: 64.7 on eight VLM benchmarks, the best among comparable-scale models; 66.03 on WorldArena, ranking first; and 93.5 on RoboTwin, the second-best average among compared action methods. These results show that the unified paradigm succeeds in preserving specialist strength while bringing understanding, reasoning, imagination, and action into one model.

\begin{center}
    \includegraphics[width=0.9\textwidth]
    {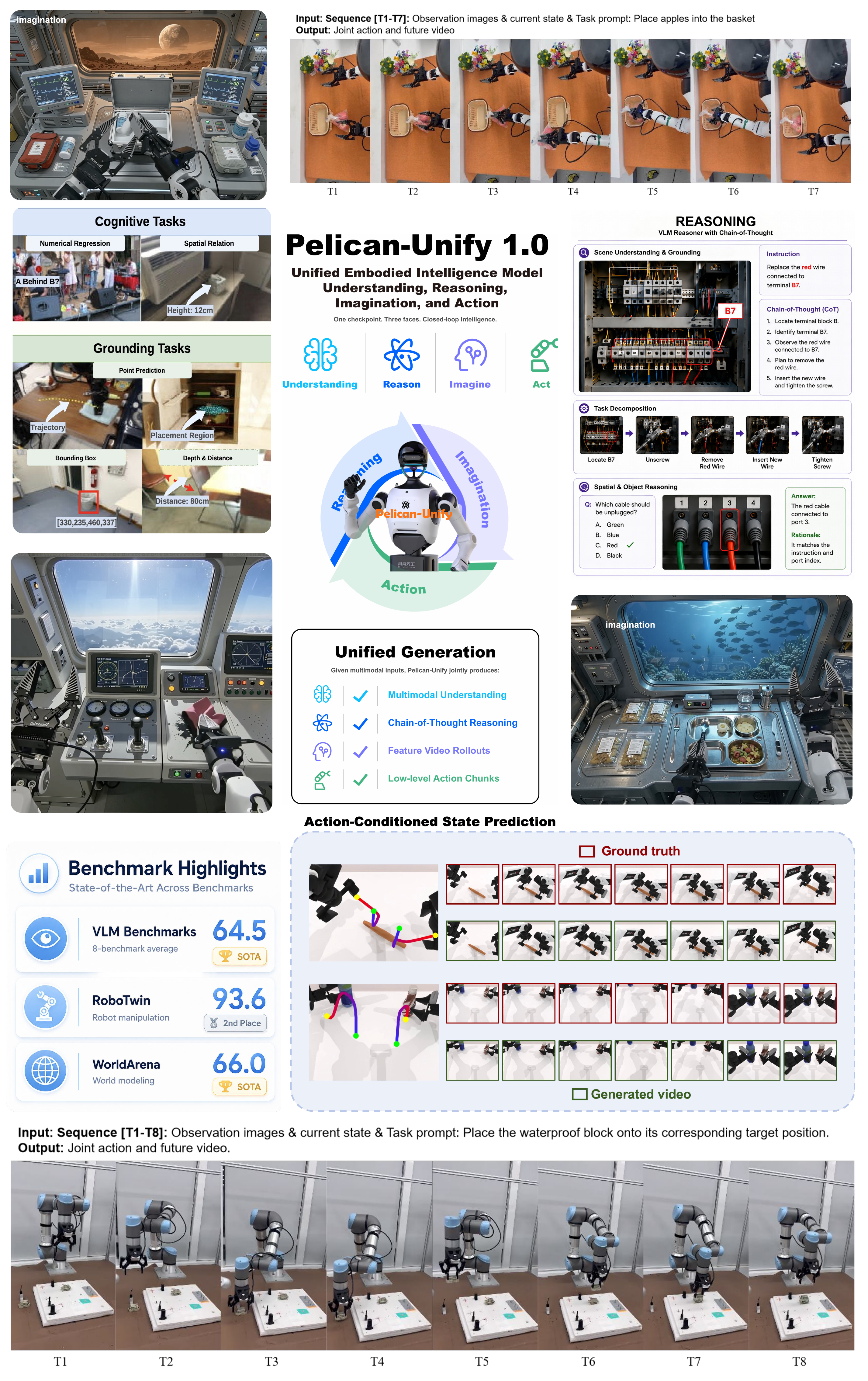}
\end{center}

\section{Introduction}

\textbf{The progress of general intelligence is reshaping embodied intelligence.}
Modern AI has moved from task-specific automation toward foundation models that can generalize across domains, follow language instructions, and reuse shared representations across many tasks. Embodied AI is undergoing a similar transition. Early robotic systems were largely built as scripted automation pipelines: perception detected predefined states, planners selected predefined routines, and controllers executed predefined motions. Recent embodied systems have become increasingly intelligent, replacing hand-designed pipelines with learned policies, vision--language grounding, large-scale imitation, and generative prediction. This shift has brought the field closer to the goal of physical intelligence: agents that can understand the world, reason about tasks, imagine what may happen, and act to change the world. Yet this progress also exposes a central question. Should physical intelligence be built by scaling separate specialists for understanding, reasoning, imagination, and action, or should these capabilities be learned as parts of one adaptive loop?

\textbf{Embodied intelligence models have advanced rapidly, but along fragmented paths.}
Vision language models such as Gemini Robotics ER~\citep{team2025gemini} and Pelican-VL~\cite{zhang2025pelican} bring strong semantic understanding and spatial-temporal reasoning to embodied settings, but they are not executable policies: they cannot directly act, cannot test their reasoning through physical consequences, and cannot ground their conclusions in the outcomes of their own behavior. Vision--language--action models such as RT-2~\citep{rt2}, $\pi_{0}$~\citep{pi0}, $\pi_{0.5}$~\citep{pi05}, OpenVLA~\citep{kim2024openvla}, and Helix~\citep{helix2025} connect language and perception to motor commands, but they typically lack explicit future imagination; as a result, their actions are often learned as imitation mappings, which limits generalization to unseen compositions, long-horizon tasks, and contact-rich interactions. World models and video generators such as Cosmos-Predict~\citep{cosmos2025}, LeWorldModel~\citep{leworldmodel}, and other recent generators~\citep{worldarena2025, bi2025motusunifiedlatentaction, wow2025} can imagine future visual states, but their imagination often remains implicit in pixels, difficult to steer with task logic, human knowledge, or language-grounded reasoning. World Action Models~\citep{wam2025} further connect imagined futures to actions, but without unified reasoning, they remain hard to interpret, hard to correct during rollout, and vulnerable to long-horizon error accumulation. The field is therefore not short of powerful components. What remains missing is a model in which understanding, reasoning, imagination, and action are learned as mutually conditioning parts of the same physical intelligence loop.

\textbf{We argue that the path forward is unification rather than modular accumulation.}
By ``unified,'' we do not mean concatenating the outputs of multiple expert networks or wiring separately trained modules into a longer pipeline. We mean that understanding, reasoning, imagination, and action should share internal representations, condition one another, and co-evolve through a common training process. This view is also consistent with embodied cognition: reasoning, imagination, and action do not operate as separable capabilities~\citep{kandel2021principles, clark2016surfing}; motor planning recruits systems involved in simulating movement~\citep{jeannerod2001neural, hesslow2002conscious}; perception is organized around what the body can do~\citep{dennett1993embodied}; and future imagination supports action selection~\citep{clark2013whatever, friston2010free}. For embodied intelligence foundation models, this principle leads to three requirements. First, \emph{understanding should be unified}: scenes, instructions, visual context, action history, and world state should be represented in a shared semantic space, reducing the semantic breaks that appear when perception, language, and action are encoded separately. Second, \emph{reasoning should be unified}: reasoning should not be a detached language trace, but a language-grounded, supervisable process whose final state directly conditions what the model imagines and executes, allowing task logic and human knowledge to guide generation. Third, \emph{generation should be unified}: future imagination and low-level action should be produced by the same conditional generative process, so actions become consequence-aware, imagination becomes task-directed, and reasoning becomes constrained by what can be simulated and executed. Under this view, the missing capability in each specialist is not an optional add-on; it is the ceiling of the other capabilities.

\textbf{We instantiate this paradigm in Pelican-Unify 1.0.}
Pelican-Unify is designed to make the understanding--reasoning--imagination--action loop a single trainable object. VLM serves as both the unified understanding engine and the unified reasoning engine: it maps scenes, instructions, visual context, and action history into a shared semantic representation, autoregressively produces a task-, action-, and future-oriented chain of thought, and projects the final hidden state into a dense latent variable \(z\). A unified Future Generator (UFG) then conditions on \(z\) and jointly generates the future video and the next chunk of low-level actions through two modality-specific output heads within the same denoising process. Language, video, and action losses all backpropagate through the shared representation. Thus, \(z\) is not a bridge between independently optimized modules; it is the learned internal state where reasoning, imagination, and action become mutually conditioning aspects of the same generative process.

\textbf{Our experiments test whether unification preserves specialist competence and improves integrated physical behavior.}
Taken apart, a single Pelican-Unify checkpoint reaches the frontiers normally occupied by separate specialists: it achieves \textbf{64.7} average on eight VLM benchmarks, among the strongest performers within models of similar scale; reaches \textbf{93.5\%} average success on RoboTwin as a competitive visuomotor policy, achieving the second-best average among compared action methods; and obtains \textbf{EWM Score 66.03} on WorldArena as a world model, ahead of dedicated world models. Taken as a whole, on a real UR5e robot and a Tienkung humanoid robot operating industrial control panels, the model achieves significant improvements on both zero-shot and compositional tasks, demonstrating the generalization and practical value of the unified architecture across different embodiments.

\newpage
\paragraph{Our contributions.}
\begin{enumerate}
    \item \textbf{A unified paradigm for embodied intelligence models.} We formulate physical intelligence as a coupled loop of understanding, reasoning, imagination, and action, and argue that the relevant unit of modeling is not an isolated specialist or a pairwise fusion, but the closed loop itself.
    \item \textbf{A concrete realization of three forms of unification.} Pelican-Unify 1.0 realizes unified understanding by building an action-oriented task state from scenes, instructions, visual context, and action history; unified reasoning by turning chain-of-thought into a dense loop state \(z\) that specifies what future should happen; and unified generation by jointly denoising future video and low-level actions from the same \(z\), with the final action read refined against the imagined future before execution.
    \item \textbf{An end-to-end objective that makes the loop learnable.} Language, video, and action supervision are trained jointly, and all three losses backpropagate through the shared latent representation. This turns reasoning, imagination, and action from inter-module messages into mutually shaping gradients within a single model.
    \item \textbf{Empirical evidence that unification preserves specialist competence and improves integrated physical behavior.} Pelican-Unified matches or exceeds dedicated models on VLM reasoning, visuomotor policy learning, and world modeling benchmarks, while achieving substantially stronger zero-shot, compositional, and long-horizon performance on real UR5e industrial control-panel manipulation tasks than the strongest modular baseline.

\end{enumerate}
\section{Pelican-Unify 1.0}

\subsection{Unified Modeling}

Pelican-Unify~1.0 fuses understanding, reasoning, and future generation into one end-to-end trainable model. A VLM backbone initialized from Qwen3-VL~\cite{Qwen3-VL} encodes all input modalities and decodes a chain-of-thought trace, while a Unified Future Generator initialized from Wan2.2~\cite{wan2025wan} jointly denoises the future video and the future action chunk. 

Formally, Pelican-Unify is a single composite map
\begin{equation}
    (\tau_t,\,\hat{v}_{t:t+H},\,\hat{a}_{t:t+H})
    =
    \mathcal{M}_{\Theta}(a_{<t},\,l,\,o_{\leq t},\,s_{\leq t}),
\end{equation}
where $\tau_t$ is the chain-of-thought trace, $\hat{v}_{t:t+H}$ the imagined future video, and $\hat{a}_{t:t+H}$ the executable action chunk. Three training signals---language, video, and action---are optimized jointly, so the shared representation is simultaneously semantic, predictive, and actionable.

\subsection{Unified Understanding}
\label{sec:understanding}

The first stage produces a single representation of \emph{what is currently happening}. The multimodal context at control step $t$ is
\begin{equation}
    c_t = (a_{<t},\, l,\, o_{\leq t},\, s_{\leq t}),
\end{equation}
which bundles the action history, language instruction, environment observations, and robot proprioceptive state. Each modality is first lifted into the VLM token space by its own embedder:
\begin{itemize}
    \item video frames $o_{\leq t}$ are encoded by a 3D video VAE $\mathcal{E}_v$;
    \item the action history $a_{<t}$ is embedded by a lightweight MLP $\mathcal{E}_a$;
    \item the language instruction $l$ is processed by the text tokenizer;
    \item the robot state $s_{\leq t}$ is mapped by a small linear projection.
\end{itemize}
We name $\mathcal{E}_v$ and $\mathcal{E}_a$ explicitly because the Unified Future Generator (Sec.~\ref{sec:future-generation}) will reuse the \emph{same} two modules to embed its diffusion inputs; understanding and generation therefore share one modality embedding space.

All tokens are concatenated and processed by the VLM to produce a unified hidden representation,
\begin{equation}
    H_t = \mathrm{VLM}_{\phi}(c_t).
\end{equation}
$H_t$ is the encoder-side state from which the VLM autoregressively decodes the chain-of-thought trace $\tau_t$ (developed in Sec.~\ref{sec:reasoning}); the actual bridge to downstream generation will be a loop state $z$ extracted at the end of $\tau_t$. Because all modalities share one transformer, perception is grounded in the instruction, language is anchored in physical context, and the action/state history is interpreted by its effect on task progress.

\subsection{Unified Reasoning}
\label{sec:reasoning}

The second stage exposes the model's reasoning in natural language. Conditioned on $H_t$, the VLM autoregressively decodes the chain-of-thought trace
\begin{equation}
    p_{\phi}(\tau_t \mid c_t)
    =
    \prod_{i=1}^{|\tau_t|}
    p_{\phi}(\tau_{t,i} \mid c_t,\,\tau_{t,<i}).
\end{equation}
The trace interleaves two complementary kinds of language:
\begin{itemize}
    \item \textbf{Video CoT} --- a description and explanation of how the scene is expected to evolve, e.g.\ which objects move, how contacts form, and how the workspace re-organizes.
    \item \textbf{Action CoT} --- an explanation and decomposition of the motor program that should realize that future, e.g.\ which sub-skill to invoke and which end-effector waypoints to target.
\end{itemize}
By putting both kinds into one sequence, the model is forced to think about ``what should happen'' and ``what should be done'' under a single causal pass.

We then summarize the trace into a dense \emph{loop state}
\begin{equation}
    z = P_{\phi}(h_{\tau_t}),
\end{equation}
where $h_{\tau_t}$ is the VLM hidden state at the end of $\tau_t$ and $P_{\phi}$ is a learned projection. $z$ is the only interface through which downstream future generation accesses the model's understanding and reasoning. Because it is shaped jointly by the language-modeling loss on $\tau_t$ and by the downstream video- and action-generation losses, $z$ must encode information that is simultaneously \emph{semantic}, \emph{predictive}, and \emph{actionable}.

\subsection{Unified Future Generation}
\label{sec:future-generation}

The third stage turns the loop state $z$ into a concrete future: the future video $v_{t:t+H}$ and the future action chunk $a_{t:t+H}$. We train the generator with continuous-time flow matching under a shared denoising backbone.

\paragraph{Shared input embedders.}
The two targets are first lifted into the generator's token space by the \emph{same} embedders used by the VLM in Sec.~\ref{sec:understanding}:
\begin{equation}
    x^v = \mathcal{E}_v(v_{t:t+H}),
    \qquad
    x^a = \mathcal{E}_a(a_{t:t+H}).
\end{equation}
At a shared diffusion time $s \sim \mathcal{U}(0,1)$, these clean latents are mixed with Gaussian noise $\epsilon^v, \epsilon^a \sim \mathcal{N}(0,I)$ to produce noisy states $x^v_s, x^a_s$; the exact mixing on the video side accommodates a prefix-conditioning mask and is detailed in Sec.~\ref{sec:diffusion}. Because $\mathcal{E}_v$ and $\mathcal{E}_a$ are tied to the VLM's input embedders, the coordinates that the generator must denoise are exactly the coordinates that the VLM has learned to read; no extra alignment module is needed between understanding and generation.

\paragraph{Shared denoising backbone.}
The two noisy streams together with the condition $z$ enter a single diffusion transformer $\mathrm{DiT}_\theta$ initialized from Wan2.2, and two lightweight heads $d_v, d_a$ read its hidden states out as per-modality velocity predictions. Writing $f_\theta$ for the DiT plus heads,
\begin{equation}
    \bigl(\hat{u}^v_s,\,\hat{u}^a_s\bigr)
    =
    f_{\theta}(x^v_s,\,x^a_s,\,z,\,s).
\end{equation}
Inside $f_\theta$, video and action tokens interact through shared self-attention, while $z$ is injected via cross-attention so that the reasoning summary continuously shapes the denoising trajectory. Modality-specific parameters appear only at the input and output boundaries---$\mathcal{E}_v, \mathcal{E}_a$ on input and $d_v, d_a$ on output; the heavy computation in between is shared across modalities.

\subsection{Conditional Diffusion over Future Video and Action}
\label{sec:diffusion}

Sec.~\ref{sec:future-generation} introduced the architecture and the loop-state condition $z$ that both streams share. We now specify the one extra condition added on the video side---the observed video prefix---and state the joint training objective. The action stream takes no further condition beyond $z$.

\paragraph{Action: plain flow matching.}
The action latent is mixed with noise in the standard way and yields the standard flow-matching target velocity,
\begin{equation}
    x^a_s = (1-s)\,x^a + s\,\epsilon^a,
    \qquad
    u^a_s = \epsilon^a - x^a.
\end{equation}

\paragraph{Video: condition on observed frames.}
We extend the video stream so that the DiT sees, alongside the noisy future latent, the observed prefix $o_{\leq t}$ encoded by the same shared $\mathcal{E}_v$. Let $M_{\mathrm{cond}}$ and $M_{\mathrm{fut}}$ be binary masks selecting the prefix and the future positions of the combined video latent, denoted $x^v$ (now covering both regions). The prefix is kept clean and only the future region is noised:
\begin{equation}
    x^v_s
    =
    M_{\mathrm{cond}} \odot x^v
    \;+\;
    M_{\mathrm{fut}} \odot
    \bigl((1-s)\,x^v + s\,\epsilon^v\bigr),
\end{equation}
with the future-only target velocity
\begin{equation}
    u^v_s = M_{\mathrm{fut}} \odot (\epsilon^v - x^v).
\end{equation}
Because the prefix is encoded by the same $\mathcal{E}_v$, the observed-frame condition lives natively in the generator's token space; no separate condition encoder is required.

\paragraph{Joint training objective.}
The video loss is computed only on the future region,
\begin{equation}
    \mathcal{L}_{\mathrm{video}}
    =
    \mathbb{E}_{s,\epsilon^v}
    \!\left[
    \bigl\| M_{\mathrm{fut}} \odot (\hat{u}^v_s - u^v_s) \bigr\|_2^2
    \right];
\end{equation}
the action loss uses a robust regression over valid action dimensions $M_a$,
\begin{equation}
    \mathcal{L}_{\mathrm{action}}
    =
    \mathbb{E}_{s,\epsilon^a}
    \!\left[
    M_a \odot \mathrm{SmoothL1}\!\left(\hat{u}^a_s,\,u^a_s\right)
    \right];
\end{equation}
and the language reasoning loss is the standard autoregressive negative log-likelihood,
\begin{equation}
    \mathcal{L}_{\mathrm{text}}
    =
    -\sum_{i=1}^{|\tau_t|}
    \log p_{\phi}(\tau_{t,i} \mid c_t,\,\tau_{t,<i}).
\end{equation}
The final objective is a weighted sum,
\begin{equation}
    \mathcal{L}
    =
    \lambda_{\mathrm{text}}\,\mathcal{L}_{\mathrm{text}}
    +
    \lambda_{\mathrm{video}}\,\mathcal{L}_{\mathrm{video}}
    +
    \lambda_{\mathrm{action}}\,\mathcal{L}_{\mathrm{action}}.
\end{equation}
All three losses flow back through the loop state $z$ \emph{and} through the shared embedders $\mathcal{E}_v, \mathcal{E}_a$, so understanding, reasoning, imagining, and acting are optimized as one closed loop rather than four stitched modules.

\section{Evaluating the Unified Model as Three Specialists}
\label{sec:simulation}

The first question for a unified model is whether unification comes at
the cost of specialist competence. If the shared loop is genuinely useful,
then training understanding, reasoning, imagination, and action together
should not weaken the model when each capability is evaluated on its own.
We therefore evaluate Pelican-Unify 1.0 in three deliberately separated
regimes: as a vision--language model on eight multimodal benchmarks, as
a visuomotor policy on the RoboTwin dual-arm simulator, and as an
action-conditioned world model on WorldArena.

\begin{figure*}[t]
    \centering
    \includegraphics[width=0.85\textwidth]{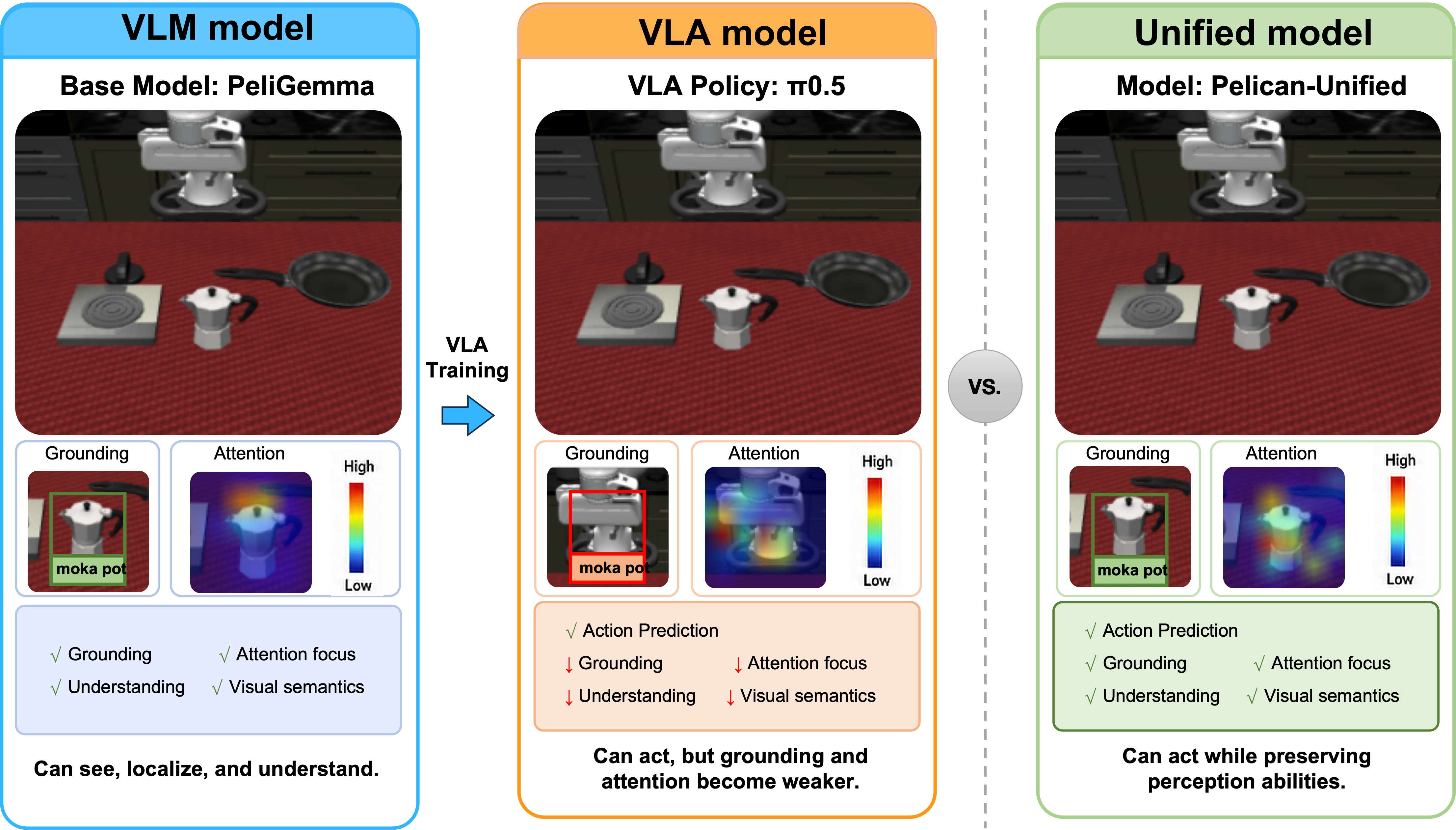}
    \caption{Starting from a base VLM, standard VLA policy training weakens grounding and attention, while Pelican-Unify 1.0 retains them and still predicts actions. Base VLM learns what and where; standard VLA weakens perception, while Pelican-Unify 1.0 preserves it and still learns what action to output.}
    \label{fig:vlm_vla_unify}
\end{figure*}

\subsection{Understanding Capability.} 

As shown in Tab.~\ref{tab:general_embodied_benchmark}, Pelican-Unify consistently achieves the highest overall performance across multimodal benchmarks for both general reasoning and embodied evaluation. Specifically, our model achieves an average score of 64.7, outperforming all compared VLA and VLM baselines. Pelican-Unify improves the overall average from 58.2, achieved by Qwen3-VL-4B-Instruct (the 4B base model we build upon) to 64.7. Furthermore, it substantially outperforms prior VLA architectures, surpassing MolmoAct (27.5) by a large margin.

A detailed breakdown of these results reveals that these gains are attained without sacrificing general multimodal capabilities. On traditional general reasoning benchmarks, Pelican-Unify matches or marginally exceeds the baseline set by Qwen3-VL-4B-Instruct. By contrast, the improvements are largest in embodied evaluation, where spatial grounding and physical understanding are critical. On embodied-oriented benchmarks such as Where2Place and PhyX, our model outperforms Qwen3-VL-4B-Instruct by $+28.2$ and $+20.6$ points, respectively. This breakdown shows that our model transfers effectively to embodied tasks without losing general reasoning: it retains strong perception while substantially improving action-aware understanding. Consequently, our unified model learns richly physically grounded representations, which provide stronger features for downstream action and video prediction.

\definecolor{pelicanrow}{RGB}{232,242,255}
\begin{table*}[t]
\centering
\small
\setlength{\tabcolsep}{4pt}
\renewcommand{\arraystretch}{1.15}

\caption{Comparison on general and embodied benchmarks. \textbf{Bold} indicates best results, \underline{underline} indicates second-best results, - indicates that the model does not possess this capability (score of 0.0).}
\label{tab:general_embodied_benchmark}

\begin{tabular}{l|ccccc|ccc|c}
\hline

\multirow{2}{*}{Method} 
& \multicolumn{5}{c|}{General Benchmark} 
& \multicolumn{3}{c|}{Embodied Benchmark} 
& \multirow{2}{*}{Avg.} \\
\cmidrule(lr){2-6} \cmidrule(lr){7-9}

& \makecell{MMMU\\\cite{yue2023mmmu}} 
& \makecell{MMBench\\\cite{liu2024mmbench}} 
& \makecell{MMStar\\\cite{chen2024we}} 
& \makecell{InfoVQA\\\cite{mathew2021infographicvqa}} 
& \makecell{ChartQA\\\cite{masry-etal-2022-chartqa}} 
& \makecell{Where2Place\\\cite{yuan2024robopointvisionlanguagemodelspatial}} 
& \makecell{PhyX\\\cite{shen2025phyxdoesmodelwits}} 
& \makecell{RefSpatial\\\cite{zhou2026roboreferspatialreferringreasoning}} 
& \\

\hline
\multicolumn{10}{l}{\textit{Vision-Language-Action Models}} \\
\hline

OpenVLA~\cite{kim2024openvla} 
& 26.3 & - & - & - & - 
& - & - & - & 3.3 \\

ECoT~\cite{zawalski2025roboticcontrolembodiedchainofthought} 
& 26.6 & 3.7 & - & - & - 
& - & 10.1 & - & 5.0 \\

MolmoAct~\cite{molmoact2025} 
& 28.4 & 55.1 & 1.2 & 41.9 & 55.9 
& 8.2 & 29.7 & - & 27.5 \\

$\pi_{0.5}$~\cite{pi05}
& 24.0 & 6.8 & 21.7 & 7.7 & 5.1 
& - & 16.2 & - & 10.2 \\

\hline
\multicolumn{10}{l}{\textit{Vision-Language Models}} \\
\hline

Gemma3-4B-IT~\cite{gemmateam2025gemma3technicalreport} 
& 39.3 & 68.6 & 37.1 & 40.9 & 50.3 
& 7.5 & 17.2 & 2.2 & 32.9 \\

Qwen3-VL-4B-Instruct~\cite{Qwen3-VL}
& \underline{52.6} & \underline{84.5} & \underline{62.9} & \underline{78.4} & \underline{81.1}
& \underline{17.0} & \underline{41.1} & \underline{48.0} & \underline{58.2} \\

\hline
\multicolumn{10}{l}{\textit{Our Unified Model}} \\
\hline

\rowcolor{pelicanrow}
Pelican-Unify 1.0
& \textbf{53.0} & \textbf{84.9} & \textbf{63.3} & \textbf{78.4} & \textbf{81.5}
& \textbf{45.2} & \textbf{61.7} & \textbf{49.3} & \textbf{64.7} \\

\hline
\end{tabular}
\end{table*}

\subsection{Action Capability.}

We evaluate Pelican-Unify on the RoboTwin 50-task dual-arm benchmark (Tab.~\ref{tab:benchmark_results}), our model achieves an average success rate of \textbf{93.5\%}. It outperforms most specialized VLA and world-model baselines, including AIM (93.1\%), LingBot-VA (92.3\%), and starVLA (88.3\%). This result demonstrates the strong manipulation capability of our model on complex tasks.

Under clean and randomized conditions, Pelican-Unify achieves 93.6\% and 93.3\% success, respectively. Per-task results show that the improvements are broad: 31 out of 50 tasks reach at least 95\% success, 39 tasks reach 90\%, and 15 tasks are solved perfectly (100\%). High-success tasks span clicking, shaking, stacking, handover, and articulated-object manipulation, indicating reliable performance across both precise contact and multi-object coordination. Failures are concentrated in the hardest long-horizon or geometry-sensitive tasks, such as hanging mugs and dustbin insertion, which require tight alignment or sustained contact. These patterns indicate that Unify training does not degrade low-level control. Rather, shared reasoning and predictive representations improve both generalizability and robustness across diverse manipulation regimes.

\definecolor{pelicanrow}{RGB}{232,242,255}
\begin{table}[t]
\small
\centering
\caption{Benchmark Results on Seen Tasks. We compare Pelican-Unify 1.0 with state-of-the-art VLA and world-model-based methods. Pelican-Unify 1.0 achieves the second-best average result while using a unified understanding--reasoning--imagination--action model. $^*$Results for X-VLA are adopted from Motus~\cite{bi2025motusunifiedlatentaction}. Unless otherwise specified, best results are highlighted in \textbf{bold}, and second-best results are \underline{underlined}. $^{\dagger}$Second-best average success rate among compared methods.}
\label{tab:benchmark_results}
\begin{tabular}{llccc}
\toprule
Type & Model & Clean & Randomized & Avg \\ \midrule
\multirow{6}{*}{\textit{VLA}}
    & $\pi_{0}$~\cite{pi0}             & 65.9 & 58.4 & 62.2 \\
    & X-VLA$^*$~\cite{zheng2026xvla}        & 72.9 & 72.8 & 72.9 \\
    & $\pi_{0.5}$~\cite{pi05}          & 82.7 & 76.8 & 79.8 \\
    & starVLA~\cite{community2026starvla} & 88.2 & 88.3 & 88.3 \\
    & ABot-M0~\cite{yang2026abot}       & 81.2 & 80.4 & 80.8 \\
    & LingBot-VLA~\cite{wu2026pragmatic} & 86.5 & 85.3 & 85.9 \\ \midrule
\multirow{7}{*}{\textit{World Model}}
    & JEPA-VLA~\cite{miao2026jepavlavideopredictiveembedding} & 73.5 & --   & --   \\
    & Motus~\cite{bi2025motusunifiedlatentaction}             & 88.7 & 87.0 & 87.9 \\
    & LingBot-VA~\cite{lingbot-va2026}                        & 92.9 & 91.6 & 92.3 \\
    & Fast-WAM~\cite{yuan2026fastwam}                         & 91.9 & 91.8 & 91.9 \\
    & Being-H0.7~\cite{beingbeyond2026beingh07}               & 90.2 & 89.6 & 89.9 \\
    & AIM~\cite{fan2026aim}                                   & \underline{94.0} & 92.1 & 93.1 \\
    & MotuBrain~\cite{motubrainteam2026motubrainadvancedworldaction} & \textbf{95.8} & \textbf{96.1} & \textbf{95.9} \\ \midrule
\multirow{1}{*}{\textit{Unified Model}}
    & \cellcolor{pelicanrow}\textbf{Pelican-Unify 1.0}$^{\dagger}$ & \cellcolor{pelicanrow}93.6 & \cellcolor{pelicanrow}\underline{93.3} & \cellcolor{pelicanrow}\underline{93.5} \\ \bottomrule
\end{tabular}
\vspace{0.2em}
\end{table}

\subsection{Imagination Capability.}
As shown in Tab.~\ref{tab:worldarena}, on the WorldArena benchmark, Pelican-Unify's imagination component achieves an EWM Score of \textbf{66.03}, ranking first. It also ranks first in 3D Accuracy (98.13) and Motion Quality (62.69), two dimensions where spatial coherence and physical plausibility are critical. On other metrics—Visual Quality (63.43), Content Consistency (60.33), Physics Adherence (61.51), Controllability (59.28)—the model performs competitively.

Automated WorldArena metrics can reward visually clean but task-irrelevant rollouts. We therefore conduct a blind human study to assess whether generated rollouts are usable for downstream control. Trained annotators rate each rollout on a $\{0,1,2\}$ scale across four axes: Controllability (preserving first-frame conditions), Task Success (achieving manipulation goals), Temporal Consistency (stable shapes without flicker), and Physical Plausibility (coherent contact and gravity). This design separates conditioning fidelity from execution quality.

As shown in Tab.~\ref{tab:worldarena_human}, Pelican-Unify 1.0 achieves the highest overall score (mean 1.76) and outperforms existing baselines. It excels on Task Success (1.81) and Controllability (2.00). The system treats first-frame conditions as manipulation goals and commits to completing them. Several video-diffusion models score near zero on Task Success despite high Temporal Consistency (e.g., Happyhorse, EnerVerse-AC), showing that visually fluent rollouts can abandon the task. 

\begin{figure*}[t]
    \centering
    \includegraphics[width=1.0\textwidth]{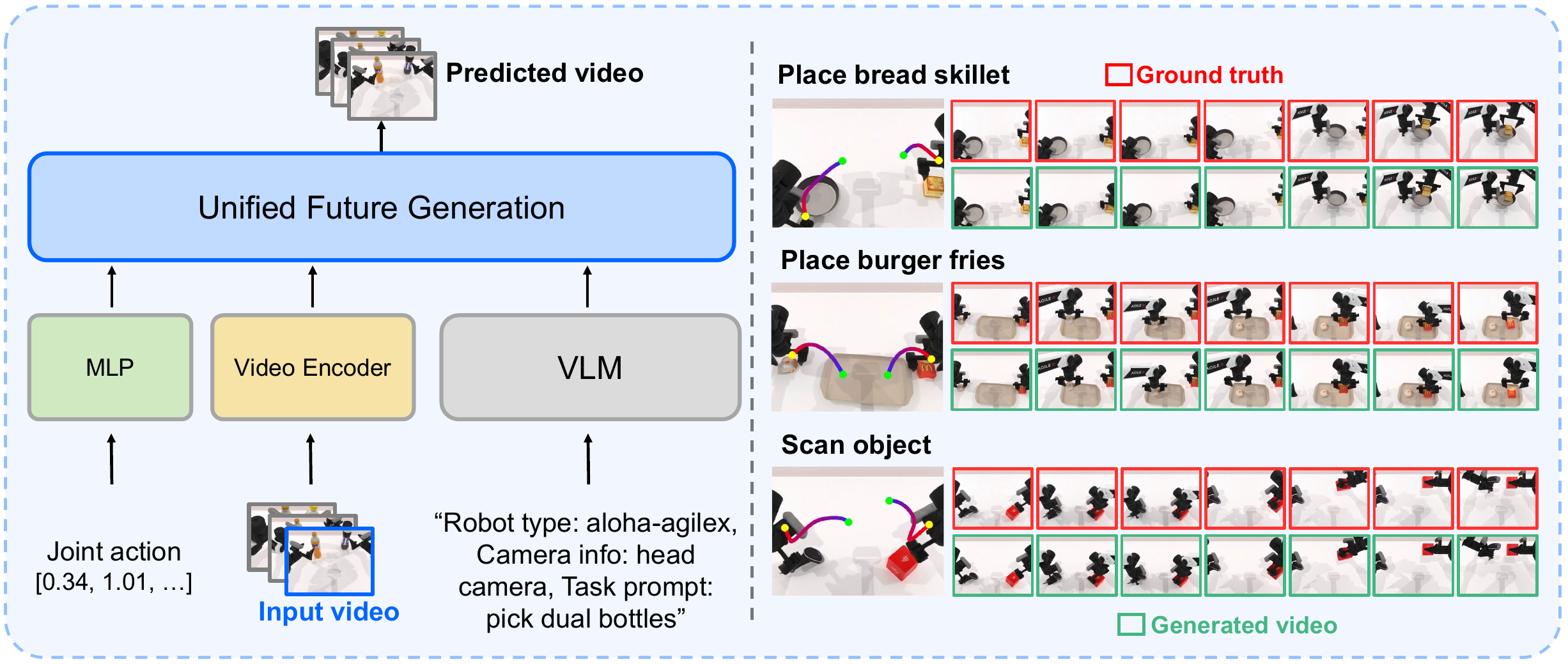}
    \caption{Pelican-Unify 1.0 can take actions as conditional inputs, enabling action-conditioned video prediction. Left: The overview of the action-conditioned video prediction model. Right: Comparison of videos generated by our method with ground truth. Our action-conditioned video prediction model achieves fine-grained alignment between input action commands and the generated video frames, given historical observations.
    }
    \label{fig:ac-pipeline}
\end{figure*}

This performance comes from the unified architecture. Unlike pure video generation models that focus on pixel-level fidelity, Pelican-Unify is pretrained on large-scale real-world robot interaction data, which encodes spatial structure and physical dynamics. The model learns 3D geometry cues—depth, object permanence, viewpoint consistency—without explicit reconstruction or physics engines. This is reflected in the 3D Accuracy. The competitive Motion Quality confirms that the model generates temporally coherent motions consistent with kinematic constraints, as these constraints are present in the pretraining action sequences. Additionally, the model supports action-conditioned video generation, improving generation quality and ensuring action-frame alignment (Fig.~\ref{fig:ac-pipeline}).

\input{worldarena_res}
\input{worldarena_human}

\section{Real-World Robot Evaluation}
\label{sec:experiments}

We ground our evaluation in real-world scenarios, as the inherent difficulty of physical manipulation exposes whether a system possesses a genuine \textbf{reason--imagine--act} closed loop or merely a chain of cooperating modules. Our experimental platform comprises a UR5e robotic arm and a Tienkung humanoid robot. The evaluation is organized around two core capabilities that a closed-loop system should satisfy:

\textbf{Compositional generalisation.} Atomic skills $A$ and $B$ compose into $A{+}B$ without any combined demonstrations.

\textbf{Zero-shot transfer.} Generalisation learned during the imagination phase transfers directly to the unified video--action model without additional training.
\subsection{Compositional Generalisation}
\label{sec:compositional}
    
\begin{figure*}[t]
    \centering
    \includegraphics[width=1\textwidth]{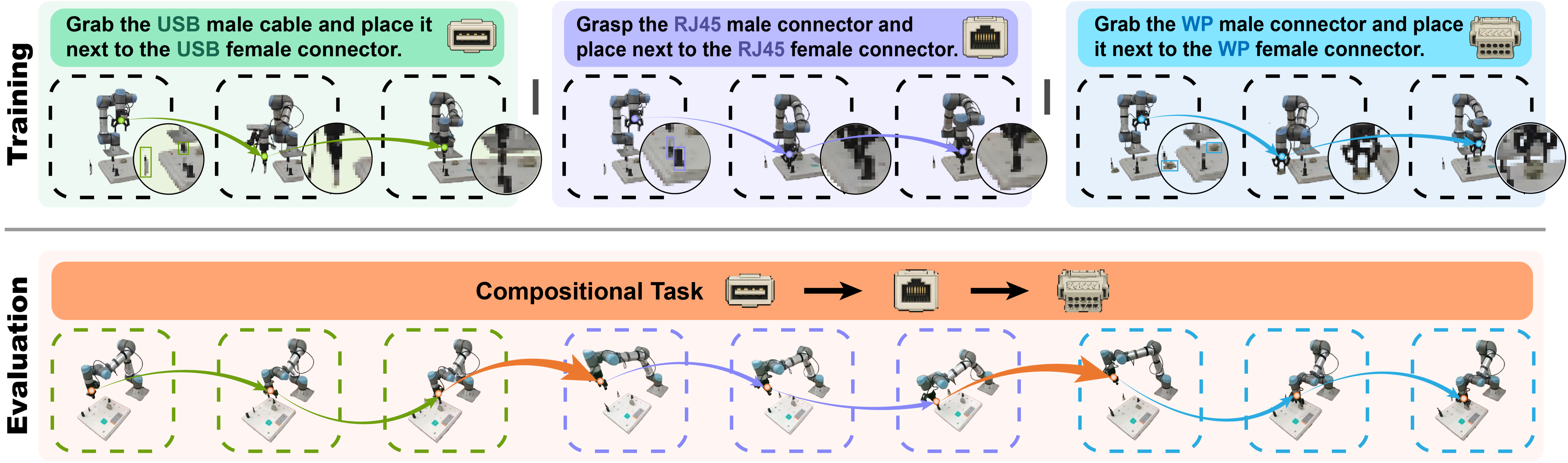}
    \caption{\textbf{Compositional generalization evaluation.} During training, the model is optimized only on atomic manipulation tasks individually, without exposure to their composed counterparts. At test time, we evaluate the model on unseen compositional tasks that require combining multiple learned skills, demonstrating strong compositional generalization ability in long-horizon embodied manipulation.}
    \label{fig:compositional}
    % \vspace{-11pt}
\end{figure*}

    We design a compositional generalisation test on the UR5e robotic arm to jointly validate two capabilities: the ability to compose unseen task combinations, and the capacity for fine-grained manipulation guided by the imagination-based world model. Specifically, the atomic tasks $\mathcal{A}$ (\textit{plug RJ45}) and $\mathcal{B}$ (\textit{waterproof}) are trained \emph{individually}, with no complete demonstration of the two chained together appearing in the training data. At test time, the robot receives a single natural-language instruction (e.g.\ \emph{``plug the RJ45 cable into port~3 and apply waterproofing''}) and must complete phase $\mathcal{A}$ followed by phase $\mathcal{B}$ within one continuous episode, as shown in Fig.~\ref{fig:compositional}.
     
    Failures are concentrated at the transition---the moment where the just-completed $A$-state must be re-perceived as the new initial condition for $B$. VLA baselines fail at this transition not because they cannot re-perceive the environment, but because their action distributions carry no representation of ``what should happen after $A$ is done''. The imagination face, having seen each atomic verb ground out into a future-frame distribution during training, can render the post-$A$ scene state and re-condition on it; the action face follows. That this succeeds without the model ever having seen a complete chained demonstration is the strongest single piece of evidence that it has learned the perception--action loop, rather than merely memorizing a richer action policy.

\begin{figure}
    \centering
    \includegraphics[width=1\linewidth]{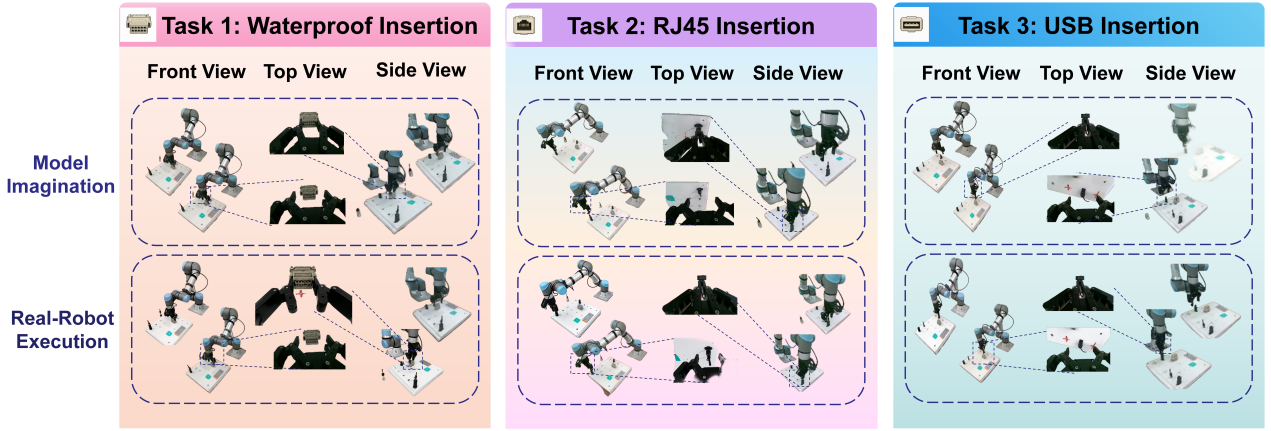}
    \caption{\textbf{Fine-grained manipulation and physical imagination capability.} Our model demonstrates strong fine-grained embodied manipulation skills in challenging connector insertion tasks, including waterproof, RJ45, and USB insertion, while also exhibiting powerful physical imagination ability to predict plausible future interactions and object dynamics under real-world constraints.}
    \label{fig:rj45_manip}
\end{figure}

    As shown in Fig.~\ref{fig:rj45_manip}, we further compare the real execution videos with the corresponding generated imagination videos during rollout. The results demonstrate that our model is able to produce physically consistent imagination and future-state predictions that closely align with the real-world observations. This indicates that the model does not merely hallucinate plausible scenes, but instead conditions its predictions on the actual environment dynamics, enabling grounded and physically coherent reasoning during execution.

\begin{figure}[!htbp]
    \centering
    \vspace{0.2em}

    \includegraphics[
        width=\linewidth,
        page=1,
        trim=6pt 0pt 6pt 0pt,
        clip,
        draft=false
    ]{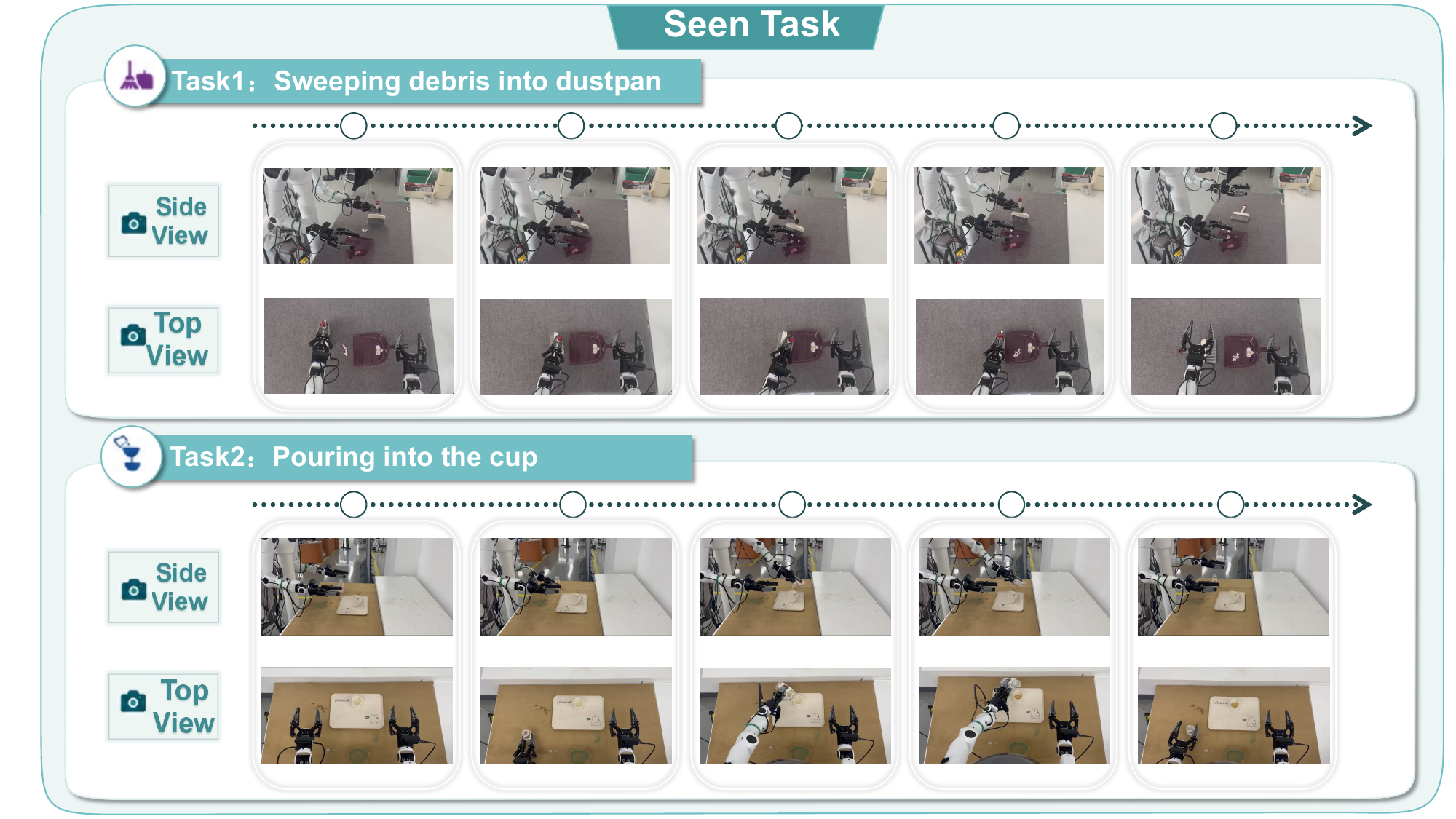}

    \vspace{0.3em}

    \includegraphics[
        width=\linewidth,
        page=2,
        trim=6pt 0pt 6pt 0pt,
        clip,
        draft=false
    ]{image_manipulation_sample_tiangun.pdf}

    \caption{
    \textbf{Execution timelines of seen and unseen robotic manipulation tasks.}
    For each task, we visualize synchronized side-view and top-view observations at five representative execution steps.
    The upper block shows two seen tasks, including sweeping debris into a dustpan and pouring into a cup, while the lower block shows an unseen cup-wiping task for evaluating cross-task generalization.
    }
    \label{fig:seen_unseen_timeline}
\end{figure}

\subsection{Zero Shot Generalization} 
\label{sec:zeroshot}

We posit that achieving robust zero shot generalization is fundamentally predicated on the broad adaptability of the underlying embodied intelligence foundation model. 
To verify this, we first conduct an exhaustive suite of evaluations across diverse cases within the Tienkung environment to establish a baseline for the model's fundamental generalization capabilities. 
Building upon this foundation, we designed a rigorous OOD benchmark within our Unified framework. 
Specifically, we performed joint training on five seen tasks, each containing an average of $300$ video-action episodes, and three unseen tasks, which were provided with only $50$ video sequences per task. 
By evaluating the task success rates in these novel scenarios, we systematically demonstrate the efficacy of our framework in bridging the domain gap and generalizing to unfamiliar environments.

As demonstrated in Tab.~\ref{tab:worldarena_human}, Pelican-Unify 1.0 attains the best overall mean score of 1.76 (rank \#1) in human evaluation, driven by the highest Task Success score of 1.81 and a perfect Controllability score of 2.00. Our unified approach significantly outperforms dedicated baselines: 0.21 ahead of the strongest video-diffusion specialist (Seedance2.0, 1.55). Notably, Pelican-Unify 1.0 is the only model to excel simultaneously in controllability, task success, temporal consistency, and physical plausibility under blind expert scrutiny, providing empirical evidence of robust generalization capabilities in embodied scenarios and establishing a solid prerequisite for subsequent experiments.
 
% \begin{figure}
%     \centering
%     \includegraphics[width=0.5\linewidth]{image.png}
%     \caption{general analysis}
%     \label{fig:general_analysis}
% \end{figure}

Building upon the robust generalization of our embodied intelligence model, the Unified framework further enhances performance through joint training. As shown in Fig.~\ref{fig:seen_unseen_timeline}, the unified model achieves strong performance across seen tasks, demonstrating the synergy between our base model and the joint training paradigm in maintaining high-fidelity execution while extending capabilities to out-of-distribution scenarios.

\section{Discussion}
\label{sec:discussion}

\subsection{What this means for the field}
\label{sec:field}

Integrated physical behavior depends on coupling, not only on
component strength. Zero-shot transfer, compositional skill use, and
long-horizon coherence are precisely the behaviors modular pipelines
have tried to engineer at the interfaces between planner, world model,
and policy. Our results suggest that these behaviors are difficult to
obtain by strengthening any one component in isolation. A policy without
future imagination remains weakly consequence-aware; a world model
without unified reasoning remains difficult to steer with task semantics
and human knowledge; and reasoning without action and imagination remains
detached from physical outcomes. What is missing in modular systems is
therefore not only more capacity, but also a training process that forces the
components to become mutually informative.

This changes what progress in embodied AI should measure. Once
understanding, reasoning, imagination, and action are trained as one
loop, improvement is not only a matter of making each specialist larger.
It also depends on how tightly the model shares representations across
modalities, how directly reasoning conditions generation, how jointly
future video and action are decoded, and how much the data itself
contains aligned observation, instruction, reasoning, action, and future
outcomes. In our ablations, the most valuable data is not merely more
data of the old form, but loop-closed data in which these signals are
annotated on the same example. Such data is valuable because a unified
model can absorb it as a coupled training signal, and because a coupled
model is precisely the kind of system that asks for it.

\subsection{Coda}
\label{sec:coda}

We began from a simple claim: physical intelligence should not be built
from fragmented capabilities. An agent that must keep improving in the
physical world needs to understand the current situation, reason about
the task, imagine possible futures, and act in ways whose consequences
feed back into the next round of understanding. Pelican-Unify~1.0 is a
concrete attempt to make this loop a single trainable object. Its
empirical signature matches the claim: the model preserves specialist
competence when evaluated one face at a time, and exhibits stronger
integrated behavior when evaluated as a whole.

We do not claim general embodied intelligence. We claim something more specific: a foundation model for embodied intelligence should allow understanding, reasoning, imagination, and action to co-evolve through a shared representation, rather than refine them as isolated systems and connect them only after training. Pelican-Unify~1.0 shows that this unification is not merely an engineering simplification. It is a practical modeling direction that preserves the strengths of specialists while enabling behaviors that depend on the loop itself. The next stage of embodied AI may therefore be shaped less by assembling larger specialists, and more by learning the shared process through which understanding, reasoning, imagination, and action become one adaptive system.

\pagebreak

\bibliography{biblio}
\bibliographystyle{abbrv}
\pagebreak
\section{Contributions}\label{contributions}
Our contributors are organized based on their roles and magnitude of contribution. The final public release will replace the group-level placeholders below with individual names after internal approval.
\subsection{Core Contributors}
Unified VLM and Action capability: Yi Zhang, Yinda Chen, Che Liu, Zeyuan Ding \\
Unified World-model capability: Jin Xu, Shilong Zou \\
Evaluation and Assets: Junwei Liao, Jiayu Hu 
\subsection{Contributors}
Xiancong Ren, Xiaopeng Zhang, Yechi Liu, Haoyuan Shi, Zecong Tang, Haosong Sun,  Renwen Cui, Kuishu Wu, Wenhai Liu, Yang Xu, Yingji Zhang, Yidong Wang, Senkang Hu, Jinpeng Lu,Nga Teng Chan, Yechen Wu, Zeting Liu, Xianzhou Hou

\subsection{Supports}
Bokai Ji, Jian Li, Yuliang Zhan
\subsection{Tech Lead}
Yong Dai
\subsection{Corresponding Authors}
Jian Tang, Xiaozhu Ju
\subsection{Acknowledgments}
We thank our families, friends, and colleagues for their patience, understanding, and unwavering support throughout this work. The pursuit of a scientific vision is never carried by the researchers alone; it is sustained by the generosity, sacrifice, and trust of those closest to us. We dedicate this paper to the memory of Jin Xu's mother, whose encouragement, even in the final moments of her life, gave strength to this work and to the people who carried it forward.
\end{document}

%% file: worldarena_res.tex
\setlength{\tabcolsep}{4pt}
\renewcommand{\arraystretch}{1.18}

\definecolor{pelicanrow}{RGB}{232,242,255}
\newcolumntype{M}{>{\raggedright\arraybackslash}p{3.4cm}}
\newcommand{\thcell}[1]{\begin{tabular}{@{}c@{}}#1\end{tabular}}

\begin{table}[t]
\centering
\caption{Performance comparison on the World Arena Benchmark (0–100 scale). Pelican-Unify achieves the best overall EWM Score (66.03, rank \#1) and demonstrates strong spatiotemporal reasoning, ranking first in both 3D Accuracy (98.13) and Motion Quality (62.69). It also delivers balanced performance across all other dimensions.}
\label{tab:worldarena}
\small
\begin{tabular}{@{}Mcccccccc@{}}
\toprule
Model &
\thcell{EWM\\Score} &
Rank &
\thcell{Visual\\Quality} &
\thcell{Motion\\Quality} &
\thcell{Content\\Consistency} &
\thcell{Physics\\Adherence} &
\thcell{3D\\Accuracy} &
Controllability \\
\midrule
\rowcolor{pelicanrow}
\textbf{Pelican-Unify} & \textbf{66.03} & \textbf{1} & 63.43 & \textbf{62.69} & 60.33 & 61.51 & \textbf{98.13} & 59.28 \\
WorldScape v0.2 & \underline{64.24} & 2 & 62.65 & 42.34 & 65.18 & \textbf{73.29} & 96.28 & 59.38 \\
FlowWAM-FiveAges & 64.12 & 3 & 63.29 & 41.05 & 66.92 & 67.82 & \underline{97.84} & 60.28 \\
MotuBrain~\cite{motubrainteam2026motubrainadvancedworldaction} & 64.07 & 4 & 60.69 & \underline{62.21} & 59.57 & 61.18 & 91.64 & 57.35 \\
FAW & 63.28 & 5 & 62.37 & 40.17 & 65.42 & \underline{69.28} & 96.85 & 58.79 \\
Goose\_Egg & 62.96 & 6 & 58.86 & 53.97 & 62.12 & 61.43 & 92.14 & 58.45 \\
ABot-PhysWorld (text)~\cite{chen2026abotphysworldinteractiveworldfoundation} & 62.63 & 7 & \textbf{64.41} & 48.34 & 63.37 & 56.73 & 85.46 & \underline{63.11} \\
Z-WM & 62.47 & 8 & \underline{64.20} & 37.43 & 64.84 & 63.88 & 96.48 & 59.80 \\
GigaWorld-1 & 62.34 & 9 & 63.04 & 39.16 & 65.17 & 64.68 & 97.02 & 57.28 \\
HeroF1 & 60.38 & 10 & 57.40 & 49.10 & 64.30 & 53.97 & 87.46 & 56.97 \\
Ctrl-World~\cite{guo2026ctrlworld} & 59.98 & 11 & 57.42 & 50.91 & 62.25 & 55.41 & 88.46 & 53.42 \\
Wan2.6~\cite{wan2026wan26} & 59.80 & 12 & 61.44 & 45.92 & 64.00 & 42.67 & 84.68 & 62.66 \\
RunWorld & 59.24 & 13 & 48.34 & 60.87 & 60.54 & 43.88 & 89.52 & 57.28 \\
CogvideoX~\cite{yang2025cogvideoxtexttovideodiffusionmodels} & 58.79 & 14 & 55.79 & 42.18 & \underline{67.71} & 50.88 & 88.28 & 55.09 \\
Veo3.1~\cite{google2026veo31} & 57.77 & 15 & 57.44 & 30.26 & \textbf{68.34} & 46.43 & 86.96 & \textbf{63.15} \\
\bottomrule
\end{tabular}
% \caption*{\footnotesize Values are reported on a 0--100 scale. Bold numbers indicate the best value in each column.}
\end{table}

%% file: worldarena_human.tex
\setlength{\tabcolsep}{6pt}
\renewcommand{\arraystretch}{1.18}

\begin{table}[t]
\centering
\caption{Human evaluation on WorldArena rollouts (0--2 scale per axis). Each rollout is independently rated along \emph{Task Success}, \emph{Controllability}, \emph{Temporal Consistency} and \emph{Physical Plausibility}; the \emph{Average} column is the unweighted mean of the four. Pelican-Unify 1.0 attains the best overall mean (1.76, rank \#1), driven by the highest Task Success (1.81) and a perfect Controllability score (2.00).}
\label{tab:worldarena_human}
\small
\begin{tabular}{@{}Mcccccc@{}}
\toprule
Model &
\thcell{Task\\Success} &
Controllability &
\thcell{Temporal\\Consistency} &
\thcell{Physical\\Plausibility} &
Average  \\
\midrule
\rowcolor{pelicanrow}
\textbf{Pelican-Unify 1.0}& \textbf{1.81} & \textbf{2.00} & \textbf{2.00} & \underline{1.23} & \textbf{1.76}  \\
% WoW                     & 1.38 & \underline{1.98} & 1.94 & \underline{1.35} & \underline{1.66} & 2 \\
Seedance2.0~\cite{seedance2026seedance20advancingvideo}~(API)            & 1.21 & 1.87 & \underline{1.98} & 1.15 & \underline{1.55}  \\
Happyhorse-1.0~\cite{happyhorse1.0i2v}~(API)     & \underline{1.65} & 1.81 & \textbf{2.00} & 0.13 & 1.40 \\
EnerVerse-AC~\cite{jiang2025enerverseacenvisioningembodiedenvironments}~          & 0.00 & 1.84 & \textbf{2.00} & \textbf{1.64} & 1.37  \\
Wan2.7~\cite{wan27official}~(API)                  & 1.19 & 1.68 & \textbf{2.00} & 0.29 & 1.29\\
Cosmos-Predict2~\cite{nvidia2025cosmosworldfoundationmodel}               & 0.63 & 1.85 & 1.79 & 0.35 & 1.16  \\
GigaWorld-0~\cite{gigaai2025gigaworld0}          & 0.33 & \underline{1.94} & \underline{1.98} & 0.13 & 1.09  \\
UnifoLM-WMA-0~\cite{unifolm-wma-0}           & 0.05 & 1.48 & \textbf{2.00} & 0.11 & 0.91  \\
\bottomrule
\end{tabular}
\end{table}

%% file: biblio.bib
@article{clark2013whatever,
  title={Whatever next? Predictive brains, situated agents, and the future of cognitive science},
  author={Clark, Andy},
  journal={Behavioral and brain sciences},
  volume={36},
  number={3},
  pages={181--204},
  year={2013},
  publisher={Cambridge University Press}
}

@article{team2025gemini,
  title={Gemini robotics: Bringing ai into the physical world},
  author={Team, Gemini Robotics and Abeyruwan, Saminda and Ainslie, Joshua and Alayrac, Jean-Baptiste and Arenas, Montserrat Gonzalez and Armstrong, Travis and Balakrishna, Ashwin and Baruch, Robert and Bauza, Maria and Blokzijl, Michiel and others},
  journal={arXiv preprint arXiv:2503.20020},
  year={2025}
}

@article{pi05,
  title={{$\pi_{0.5}$}: a Vision-Language-Action Model with Open-World Generalization},
  author={Intelligence, Physical and Black, Kevin and Brown, Noah and Darpinian, James and Dhabalia, Karan and Driess, Danny and Esmail, Adnan and Equi, Michael and Finn, Chelsea and Fusai, Niccolo and others},
  journal={arXiv preprint arXiv:2504.16054},
  year={2025}
}

@article{leworldmodel,
  title={Leworldmodel: Stable end-to-end joint-embedding predictive architecture from pixels},
  author={Maes, Lucas and Lidec, Quentin Le and Scieur, Damien and LeCun, Yann and Balestriero, Randall},
  journal={arXiv preprint arXiv:2603.19312},
  year={2026}
}

@inproceedings{rt2,
  title={Rt-2: Vision-language-action models transfer web knowledge to robotic control},
  author={Zitkovich, Brianna and Yu, Tianhe and Xu, Sichun and Xu, Peng and Xiao, Ted and Xia, Fei and Wu, Jialin and Wohlhart, Paul and Welker, Stefan and Wahid, Ayzaan and others},
  booktitle={Conference on Robot Learning},
  pages={2165--2183},
  year={2023},
  organization={PMLR}
}

@article{pi0,
  title   = {{$\pi_0$}: A Vision-Language-Action Flow Model for General Robot Control},
  author={Black, Kevin and Brown, Noah and Driess, Danny and Esmail, Adnan and Equi, Michael and Finn, Chelsea and Fusai, Niccolo and Groom, Lachy and Hausman, Karol and Ichter, Brian and others},
  journal={arXiv preprint arXiv:2410.24164},
  year={2024}
}

@misc{helix2025,
  title={Helix: A vision-language-action model for generalist humanoid control},
  author={Figure, AI},
  journal={Figure AI News},
  year={2024}
}

@article{cosmos2025,
  title={World simulation with video foundation models for physical ai},
  author={Ali, Arslan and Bai, Junjie and Bala, Maciej and Balaji, Yogesh and Blakeman, Aaron and Cai, Tiffany and Cao, Jiaxin and Cao, Tianshi and Cha, Elizabeth and Chao, Yu-Wei and others},
  journal={arXiv preprint arXiv:2511.00062},
  year={2025}
}

@article{worldarena2025,
  title={WorldArena: A Unified Benchmark for Evaluating Perception and Functional Utility of Embodied World Models},
  author={Shang, Yu and Li, Zhuohang and Ma, Yiding and Su, Weikang and Jin, Xin and Wang, Ziyou and Jin, Lei and Zhang, Xin and Tang, Yinzhou and Su, Haisheng and others},
  journal={arXiv preprint arXiv:2602.08971},
  year={2026}
}

@article{wow2025,
  title={Wow: Towards a world omniscient world model through embodied interaction},
  author={Chi, Xiaowei and Jia, Peidong and Fan, Chun-Kai and Ju, Xiaozhu and Mi, Weishi and Zhang, Kevin and Qin, Zhiyuan and Tian, Wanxin and Ge, Kuangzhi and Li, Hao and others},
  journal={arXiv preprint arXiv:2509.22642},
  year={2025}
}

@article{wam2025,
  title={World action models are zero-shot policies},
  author={Ye, Seonghyeon and Ge, Yunhao and Zheng, Kaiyuan and Gao, Shenyuan and Yu, Sihyun and Kurian, George and Indupuru, Suneel and Tan, You Liang and Zhu, Chuning and Xiang, Jiannan and others},
  journal={arXiv preprint arXiv:2602.15922},
  year={2026}
}

@article{kim2024openvla,
  title={Openvla: An open-source vision-language-action model},
  author={Kim, Moo Jin and Pertsch, Karl and Karamcheti, Siddharth and Xiao, Ted and Balakrishna, Ashwin and Nair, Suraj and Rafailov, Rafael and Foster, Ethan and Lam, Grace and Sanketi, Pannag and others},
  journal={arXiv preprint arXiv:2406.09246},
  year={2024}
}

@misc{shen2025phyxdoesmodelwits,
title={PhyX: Does Your Model Have the "Wits" for Physical Reasoning?},
author={Hui Shen and Taiqiang Wu and Qi Han and Yunta Hsieh and Jizhou Wang and Yuyue Zhang and Yuxin Cheng and Zijian Hao and Yuansheng Ni and Xin Wang and Zhongwei Wan and Kai Zhang and Wendong Xu and Jing Xiong and Ping Luo and Wenhu Chen and Chaofan Tao and Zhuoqing Mao and Ngai Wong},
year={2025},
eprint={2505.15929},
archivePrefix={arXiv},
primaryClass={cs.AI},
url={https://arxiv.org/abs/2505.15929},
}

@article{lingbot-va2026,
  title={Causal World Modeling for Robot Control},
  author={Li, Lin and Zhang, Qihang and Luo, Yiming and Yang, Shuai and Wang, Ruilin and Han, Fei and Yu, Mingrui and Gao, Zelin and Xue, Nan and Zhu, Xing and Shen, Yujun and Xu, Yinghao},
  journal={arXiv preprint arXiv:2601.21998},
  year={2026}
}

@misc{bi2025motusunifiedlatentaction,
      title={Motus: A Unified Latent Action World Model}, 
      author={Hongzhe Bi and Hengkai Tan and Shenghao Xie and Zeyuan Wang and Shuhe Huang and Haitian Liu and Ruowen Zhao and Yao Feng and Chendong Xiang and Yinze Rong and Hongyan Zhao and Hanyu Liu and Zhizhong Su and Lei Ma and Hang Su and Jun Zhu},
      year={2025},
      eprint={2512.13030},
      archivePrefix={arXiv},
      primaryClass={cs.CV},
      url={https://arxiv.org/abs/2512.13030}, 
}

@book{kandel2021principles,
  title     = {Principles of Neural Science},
  author    = {Kandel, Eric R. and Koester, John D. and Mack, Sarah H. and Siegelbaum, Steven A.},
  edition   = {6},
  year      = {2021},
  publisher = {McGraw-Hill Education},
  address   = {New York}
}

@article{jeannerod2001neural,
  author  = {Jeannerod, Marc},
  title   = {Neural Simulation of Action: A Unifying Mechanism for Motor Cognition},
  journal = {NeuroImage},
  volume  = {14},
  number  = {1},
  pages   = {S103--S109},
  year    = {2001}
}

@article{hesslow2002conscious,
  author  = {Hesslow, Germund},
  title   = {Conscious Thought as Simulation of Behaviour and Perception},
  journal = {Trends in Cognitive Sciences},
  volume  = {6},
  number  = {6},
  pages   = {242--247},
  year    = {2002}
}

@article{friston2010free,
  author  = {Friston, Karl},
  title   = {The Free-Energy Principle: A Unified Brain Theory?},
  journal = {Nature Reviews Neuroscience},
  volume  = {11},
  number  = {2},
  pages   = {127--138},
  year    = {2010}
}

@book{clark2016surfing,
  title     = {Surfing Uncertainty: Prediction, Action, and the Embodied Mind},
  author    = {Clark, Andy},
  year      = {2016},
  publisher = {Oxford University Press},
  address   = {Oxford}
}

@misc{dennett1993embodied,
  title={The embodied mind: Cognitive science and human experience},
  author={Dennett, Daniel C},
  year={1993},
  publisher={JSTOR}
}

@article{community2026starvla,
  title={StarVLA: A Lego-like Codebase for Vision-Language-Action Model Developing},
  author={Community, StarVLA},
  journal={arXiv preprint arXiv:2604.05014},
  year={2026}
}

@article{yang2026abot,
  title={Abot-m0: Vla foundation model for robotic manipulation with action manifold learning},
  author={Yang, Yandan and Zeng, Shuang and Lin, Tong and Chang, Xinyuan and Qi, Dekang and Xiao, Junjin and Liu, Haoyun and Chen, Ronghan and Chen, Yuzhi and Huo, Dongjie and others},
  journal={arXiv preprint arXiv:2602.11236},
  year={2026}
}

@article{wu2026pragmatic,
  title={A Pragmatic VLA Foundation Model},
  author={Wu, Wei and Lu, Fan and Wang, Yunnan and Yang, Shuai and Liu, Shi and Wang, Fangjing and Zhu, Qian and Sun, He and Wang, Yong and Ma, Shuailei and others},
  journal={arXiv preprint arXiv:2601.18692},
  year={2026}
}

@misc{molmoact2025,
      title={MolmoAct: Action Reasoning Models that can Reason in Space}, 
      author={Jason Lee and Jiafei Duan and Haoquan Fang and Yuquan Deng and Shuo Liu and Boyang Li and Bohan Fang and Jieyu Zhang and Yi Ru Wang and Sangho Lee and Winson Han and Wilbert Pumacay and Angelica Wu and Rose Hendrix and Karen Farley and Eli VanderBilt and Ali Farhadi and Dieter Fox and Ranjay Krishna},
      year={2025},
      eprint={2508.07917},
      archivePrefix={arXiv},
      primaryClass={cs.RO},
      url={https://arxiv.org/abs/2508.07917}
}

@misc{miao2026jepavlavideopredictiveembedding,
      title={JEPA-VLA: Video Predictive Embedding is Needed for VLA Models}, 
      author={Shangchen Miao and Ningya Feng and Jialong Wu and Ye Lin and Xu He and Dong Li and Mingsheng Long},
      year={2026},
      eprint={2602.11832},
      archivePrefix={arXiv},
      primaryClass={cs.CV},
      url={https://arxiv.org/abs/2602.11832}, 
}

@misc{gemmateam2025gemma3technicalreport,
  title={Gemma 3 technical report},
  author={Kamath, Aishwarya and Ferret, Johan and Pathak, Shreya and Vieillard, Nino and Merhej, Ramona and Perrin, Sarah and Matejovicova, Tatiana and Ram{\'e}, Alexandre and Rivi{\`e}re, Morgane and Rouillard, Louis and others},
  journal={arXiv preprint arXiv:2503.19786},
  volume={4},
  year={2025},
  publisher={ArXiv}
}

@article{Qwen3-VL,
  title={Qwen3-VL Technical Report},
  author={Bai, Shuai and others},
  journal={arXiv preprint arXiv:2511.21631},
  year={2025}
}

@article{zhang2025pelican,
  title={Pelican-VL 1.0: A Foundation Brain Model for Embodied Intelligence},
  author={Zhang, Yi and Liu, Che and Ren, Xiancong and Ni, Hanchu and Zhang, Shuai and Ding, Zeyuan and Hu, Jiayu and Shan, Hanzhe and Niu, Zhenwei and Liu, Zhaoyang and others},
  journal={arXiv preprint arXiv:2511.00108},
  year={2025}
}

@article{wan2025wan,
  title={Wan: Open and advanced large-scale video generative models},
  author={Wan, Team and Wang, Ang and Ai, Baole and Wen, Bin and Mao, Chaojie and Xie, Chen-Wei and Chen, Di and Yu, Feiwu and Zhao, Haiming and Yang, Jianxiao and others},
  journal={arXiv preprint arXiv:2503.20314},
  year={2025}
}

@article{yuan2026fastwam,
  title={Fast-WAM: Do World Action Models Need Test-time Future Imagination?},
  author={Tianyuan Yuan and Zibin Dong and Yicheng Liu and Hang Zhao},
  journal={arXiv preprint arXiv:2603.16666},
  year={2026},
  url={https://arxiv.org/abs/2603.16666}
}

@article{beingbeyond2026beingh07,
  title={Being-H0.7: A Latent World-Action Model from Egocentric Videos},
  author={Luo, Hao and Zhang, Wanpeng and Feng, Yicheng and Zheng, Sipeng and Xu, Haiweng and Xu, Chaoyi and Xi, Ziheng and Fu, Yuhui and Lu, Zongqing},
  journal={arXiv preprint arXiv:2605.00078},
  year={2026}
}

@inproceedings{yue2023mmmu,
  title={MMMU: A Massive Multi-discipline Multimodal Understanding and Reasoning Benchmark for Expert AGI},
  author={Xiang Yue and Yuansheng Ni and Kai Zhang and Tianyu Zheng and Ruoqi Liu and Ge Zhang and Samuel Stevens and Dongfu Jiang and Weiming Ren and Yuxuan Sun and Cong Wei and Botao Yu and Ruibin Yuan and Renliang Sun and Ming Yin and Boyuan Zheng and Zhenzhu Yang and Yibo Liu and Wenhao Huang and Huan Sun and Yu Su and Wenhu Chen},
  booktitle={Proceedings of CVPR},
  year={2024},
}

@article{fan2026aim,
  title={AIM: Intent-Aware Unified world action Modeling with Spatial Value Maps},
  author={Fan, Liaoyuan and Xu, Zetian and Cao, Chen and Zhang, Wenyao and Yuan, Mingqi and Chen, Jiayu},
  journal={arXiv preprint arXiv:2604.11135},
  year={2026}
}

@inproceedings{liu2024mmbench,
  title={Mmbench: Is your multi-modal model an all-around player?},
  author={Liu, Yuan and Duan, Haodong and Zhang, Yuanhan and Li, Bo and Zhang, Songyang and Zhao, Wangbo and Yuan, Yike and Wang, Jiaqi and He, Conghui and Liu, Ziwei and others},
  booktitle={European conference on computer vision},
  pages={216--233},
  year={2024},
  organization={Springer}
}

@inproceedings{zheng2026xvla,
  title={X-VLA: Soft-Prompted Transformer as Scalable Cross-Embodiment Vision-Language-Action Model},
  author={Zheng, Jinliang and Li, Jianxiong and Wang, Zhihao and Liu, Dongxiu and Kang, Xirui and Feng, Yuchun and Zheng, Yinan and Zou, Jiayin and Chen, Yilun and Zeng, Jia and Wang, Tai and Zhang, Ya-Qin and Liu, Jingjing and Zhan, Xianyuan},
  booktitle={The Fourteenth International Conference on Learning Representations (ICLR)},
  year={2026},
  url={https://iclr.cc/virtual/2026/poster/10007740}
}

@article{chen2024we,
  title={Are We on the Right Way for Evaluating Large Vision-Language Models?},
  author={Chen, Lin and Li, Jinsong and Dong, Xiaoyi and Zhang, Pan and Zang, Yuhang and Chen, Zehui and Duan, Haodong and Wang, Jiaqi and Qiao, Yu and Lin, Dahua and others},
  journal={arXiv preprint arXiv:2403.20330},
  year={2024}
}

@misc{mathew2021infographicvqa,
      title={InfographicVQA}, 
      author={Minesh Mathew and Viraj Bagal and Rubèn Pérez Tito and Dimosthenis Karatzas and Ernest Valveny and C. V Jawahar},
      year={2021},
      eprint={2104.12756},
      archivePrefix={arXiv},
      primaryClass={cs.CV},
      url={https://arxiv.org/abs/2104.12756}, 
}

@inproceedings{masry-etal-2022-chartqa,
    title = "{C}hart{QA}: A Benchmark for Question Answering about Charts with Visual and Logical Reasoning",
    author = "Masry, Ahmed  and
      Long, Do  and
      Tan, Jia Qing  and
      Joty, Shafiq  and
      Hoque, Enamul",
    booktitle = "Findings of the Association for Computational Linguistics: ACL 2022",
    month = may,
    year = "2022",
    address = "Dublin, Ireland",
    publisher = "Association for Computational Linguistics",
    url = "https://aclanthology.org/2022.findings-acl.177",
    doi = "10.18653/v1/2022.findings-acl.177",
    pages = "2263--2279",
}

@misc{yuan2024robopointvisionlanguagemodelspatial,
      title={RoboPoint: A Vision-Language Model for Spatial Affordance Prediction for Robotics}, 
      author={Wentao Yuan and Jiafei Duan and Valts Blukis and Wilbert Pumacay and Ranjay Krishna and Adithyavairavan Murali and Arsalan Mousavian and Dieter Fox},
      year={2024},
      eprint={2406.10721},
      archivePrefix={arXiv},
      primaryClass={cs.RO},
      url={https://arxiv.org/abs/2406.10721}, 
}

@misc{motubrainteam2026motubrainadvancedworldaction,
      title={MotuBrain: An Advanced World Action Model for Robot Control}, 
      author={MotuBrain Team and Chendong Xiang and Fan Bao and Haitian Liu and Hengkai Tan and Hongzhe Bi and James Li and Jiabao Liu and Jingrui Pang and Kiro Jing and Louis Liu and Mengchen Cai and Rongxu Cui and Ruowen Zhao and Runqing Wang and Shuhe Huang and Yao Feng and Yinze Rong and Zeyuan Wang and Jun Zhu},
      year={2026},
      eprint={2604.27792},
      archivePrefix={arXiv},
      primaryClass={cs.RO},
      url={https://arxiv.org/abs/2604.27792}, 
}

@misc{zhou2026roboreferspatialreferringreasoning,
      title={RoboRefer: Towards Spatial Referring with Reasoning in Vision-Language Models for Robotics}, 
      author={Enshen Zhou and Jingkun An and Cheng Chi and Yi Han and Shanyu Rong and Chi Zhang and Pengwei Wang and Zhongyuan Wang and Tiejun Huang and Lu Sheng and Shanghang Zhang},
      year={2026},
      eprint={2506.04308},
      archivePrefix={arXiv},
      primaryClass={cs.RO},
      url={https://arxiv.org/abs/2506.04308}, 
}

@misc{chen2026abotphysworldinteractiveworldfoundation,
      title={ABot-PhysWorld: Interactive World Foundation Model for Robotic Manipulation with Physics Alignment}, 
      author={Yuzhi Chen and Ronghan Chen and Dongjie Huo and Yandan Yang and Dekang Qi and Haoyun Liu and Tong Lin and Shuang Zeng and Junjin Xiao and Xinyuan Chang and Feng Xiong and Xing Wei and Zhiheng Ma and Mu Xu},
      year={2026},
      eprint={2603.23376},
      archivePrefix={arXiv},
      primaryClass={cs.CV},
      url={https://arxiv.org/abs/2603.23376}, 
}

@inproceedings{guo2026ctrlworld,
  title={Ctrl-World: A Controllable Generative World Model for Robot Manipulation},
  author={Guo, Yanjiang and Shi, Lucy Xiaoyang and Chen, Jianyu and Finn, Chelsea},
  booktitle={The Fourteenth International Conference on Learning Representations (ICLR)},
  year={2026},
  url={https://iclr.cc/virtual/2026/poster/10011332}
}

@misc{wan2026wan26,
  author = {Wan Team},
  title = {Wan2.6: A State-of-the-art Video Generation Model},
  year = {2026},
  howpublished = {\url{Wan AI: Leading AI Video Generation Model}}, 
  note = {Accessed: 2026-05-14}
}

@misc{yang2025cogvideoxtexttovideodiffusionmodels,
      title={CogVideoX: Text-to-Video Diffusion Models with An Expert Transformer}, 
      author={Zhuoyi Yang and Jiayan Teng and Wendi Zheng and Ming Ding and Shiyu Huang and Jiazheng Xu and Yuanming Yang and Wenyi Hong and Xiaohan Zhang and Guanyu Feng and Da Yin and Yuxuan Zhang and Weihan Wang and Yean Cheng and Bin Xu and Xiaotao Gu and Yuxiao Dong and Jie Tang},
      year={2025},
      eprint={2408.06072},
      archivePrefix={arXiv},
      primaryClass={cs.CV},
      url={https://arxiv.org/abs/2408.06072}, 
}

@misc{google2026veo31,
  author = {Google DeepMind},
  title = {Veo 3.1: Our most capable generative video model},
  year = {2026},
  howpublished = {\url{https://deepmind.google/technologies/veo/}}, 
  note = {Accessed: 2026-05-14}
}

@misc{seedance2026seedance20advancingvideo,
       title={Seedance 2.0: Advancing video generation for world complexity},
  author={Seedance, Team and Chen, De and Chen, Liyang and Chen, Xin and Chen, Ying and Chen, Zhuo and Chen, Zhuowei and Cheng, Feng and Cheng, Tianheng and Cheng, Yufeng and others},
  journal={arXiv preprint arXiv:2604.14148},
  year={2026}
}

@misc{jiang2025enerverseacenvisioningembodiedenvironments,
      title={EnerVerse-AC: Envisioning Embodied Environments with Action Condition}, 
      author={Yuxin Jiang and Shengcong Chen and Siyuan Huang and Liliang Chen and Pengfei Zhou and Yue Liao and Xindong He and Chiming Liu and Hongsheng Li and Maoqing Yao and Guanghui Ren},
      year={2025},
      eprint={2505.09723},
      archivePrefix={arXiv},
      primaryClass={cs.RO},
      url={https://arxiv.org/abs/2505.09723}, 
}

@online{happyhorse1.0i2v,
  author = {Happyhorse Team},
  title = {Happyhorse-1.0},
  year = {2026},
  url = {https://www.happyhorse.com/},
  urldate = {2026-05-14},
  organization = {Happyhorse}
}

@online{wan27official,
  author = {Wan Team},
  title = {Wan2.7},
  year = {2026},
  url = {https://wan2-7.io/},
  urldate = {2026-05-14},
  organization = {Wan}
}

@misc{nvidia2025cosmosworldfoundationmodel,
      title={Cosmos world foundation model platform for physical ai},
  author={Agarwal, Niket and Ali, Arslan and Bala, Maciej and Balaji, Yogesh and Barker, Erik and Cai, Tiffany and Chattopadhyay, Prithvijit and Chen, Yongxin and Cui, Yin and Ding, Yifan and others},
  journal={arXiv preprint arXiv:2501.03575},
  year={2025}
}

@misc{gigaai2025gigaworld0,
      title={GigaWorld-0: World Models as Data Engine to Empower Embodied AI},
      author={GigaAI},
      year={2025},
      eprint={2511.19861},
      archivePrefix={arXiv},
      primaryClass={cs.CV},
      url={https://arxiv.org/abs/2511.19861},
}

@misc{unifolm-wma-0,
  author       = {Unitree},
  title        = {UnifoLM-WMA-0: A World-Model-Action (WMA) Framework under UnifoLM Family},
  year         = {2025},
}

@misc{zawalski2025roboticcontrolembodiedchainofthought,
      title={Robotic Control via Embodied Chain-of-Thought Reasoning}, 
      author={Michał Zawalski and William Chen and Karl Pertsch and Oier Mees and Chelsea Finn and Sergey Levine},
      year={2025},
      eprint={2407.08693},
      archivePrefix={arXiv},
      primaryClass={cs.RO},
      url={https://arxiv.org/abs/2407.08693}, 
}
